\documentclass{article}

\PassOptionsToPackage{numbers,compress}{natbib}
\usepackage[final,dblblindworkshop]{neurips_2026}

\usepackage[utf8]{inputenc}
\usepackage[T1]{fontenc}
\usepackage[hidelinks]{hyperref}
\usepackage{url}
\usepackage{booktabs}
\usepackage{amsmath}
\usepackage{amsfonts}
\usepackage{microtype}
\usepackage{xcolor}
\usepackage{algorithm}
\usepackage{algpseudocode}

\newcommand{\method}{\textsc{GradaSlim}}

\newcommand{\energy}{\mathcal{E}}

\title{From DNA Design to DNA Slimming: Auditable Agentic Discovery of a Deletion-Only Designer}
\workshoptitle{Agentic AI for Biological Discovery}

\author{%
  Joel Shor \\
  AI Bio Design at the Allen Institute \& Move37 Labs\\
  \texttt{joel.shor@alleninstitute.org}
}

\begin{document}

\maketitle

\begin{abstract}
Compact regulatory DNA can free up space in vector payloads, reduce synthesis and assay burden, and expose which sequence features drive predicted activity.  Yet most model-based nucleic-acid designers optimize fixed-length sequences through
substitutions; they do not ask which bases of an existing functional element can be removed while retaining predicted activity.  We define the task of \emph{sequence slimming} as selecting an exact-length, order-preserving subsequence while retaining activity.  Modeled on the design benchmark NucleoBench, we propose a quantitative evaluation for slimming that balances sequence reduction with maintaining function. Each slimmer must return both the subsequence and its source indices, which can be used to verify that the slimmer obeyed task requirements. To our knowledge, this is the first dedicated benchmark of this deletion-only problem.
The coding agent Empirical Research Assistant (ERA) then searched over executable designer programs. ERA received the task prompt and a successful substitution-only designer GrAdaBeam as a starting program, and it modified the designer to produce \method{}.
We report held-out evaluations for five transcription-factor binding targets, comparing random, greedy, and ERA-guided slimming at 400 and 100 bp. ERA has the highest mean in 9/10 settings. Paired bootstrap intervals for ERA minus greedy are above zero in all five 400-bp settings, below zero in one 100-bp setting, and overlap zero in the remaining four.
\end{abstract}

\section{Introduction}

Regulatory elements that are compact have practical benefits. Viral vectors are limited in the length of the payload they can deliver, and more compact payloads can show improved properties, such as transducibility
\citep{psatha2025compact}.  However, natural flanking sequences sometimes contain desirable properties~\citep{lopezrivera2020buffered}, making the slimming problem a complex one.  Slimming therefore needs to be considered as a size/function tradeoff.

Model-based DNA designers take a deep sequence-to-function model (oracle) and iteratively make substitutions to maximize the oracle-predicted activity. When the oracles are trained to predict transcription-factor binding and motif syntax, such as BPNet~\citep{avsec2021bpnet}, this iterative approach designs for regulatory behavior. This technique has been validated biologically~\citep{taskiran2024celltype,gosai2024machineguided}. Only recently has there been a quantitative comparison of model-based, substitution-only designers~\citep{shor2026gradabeam}. This comparison explored a number of substitution designers~\citep{linder2021fast,schreiber2020ledidi,sinai2020adalead,shor2026gradabeam}, but no deletion-based designers.

GrAdaBeam is a particularly relevant starting point for sequence slimming because
  it was designed as a hybrid of gradient guidance and adaptive discrete search
  \citep{shor2026gradabeam}. It operates over proposed discrete actions. In principle, the
  action space can therefore simply be changed from substitutions to deletions.
  The gradient-based components do not transfer as directly. For example, a substitution modifies one
  position in a fixed-length representation, whereas deleting a base shifts every
  downstream coordinate and creates a new sequence junction. Consequently, the
  substitution gradients used by GrAdaBeam, Ledidi, and FastSeqProp cannot be
  interpreted directly as deletion gradients
  \citep{schreiber2020ledidi,linder2021fast}. Purely gradient-based designers would
  require a new variable-length relaxation or alignment-aware parameterization,
  whereas GrAdaBeam provides a natural discrete search architecture in which to test
  deletion actions.

Previous work has explored short enhancer design in the wet lab~\citep{lopezrivera2020buffered}. Earlier computational enhancer design used fixed-length substitutions or mixed substitution, insertion, and deletion paths \citep{martinez2013synthetic}.  To our knowledge, prior work has not defined and benchmarked the dedicated problem studied here: return an \emph{exact-length, deletion-only, order-preserving subsequence} of an existing functional regulatory sequence.

Agentic program search such as FunSearch, AlphaEvolve, and the Empirical Research Assistant (ERA) use language models plus automated evaluation to
search over executable programs \citep{romeraparedes2024funsearch,
novikov2025alphaevolve,aygun2026era}.  In our setting, ERA searches over \emph{executable designer programs}, while each proposed program searches over \emph{DNA subsequences}.  This nested problem decomposition is attractive because the output of the outer loop is ordinary code that can be inspected, versioned, and tested directly on biological tasks.

This short paper makes three contributions.  First, we define a machine-verifiable
evaluation framework for deletion-only slimming.  Second, we describe an algorithm discovered by ERA in a deletion-only sequence-slimming task: a two-phase search that learns a salience map from scored discrete candidates.  Third, we report five held-out BPNet studies comparing the discovered program with random and greedy controls under a common, target-separated protocol. The full algorithm appears in Appendix~\ref{app:algorithm}.

\section{Defining sequence slimming}

\paragraph{Problem definition.}
  Given a source DNA sequence $x$ of length $L$ and a target length $m<L$, a
  slimmer must return two objects: a candidate sequence $\widehat{x}$ and its
  claimed retained source positions $I=(i_1,\ldots,i_m)$. An output is valid only
  if both objects have length $m$,
  $1\leq i_1<\cdots<i_m\leq L$, and
  $\widehat{x}_j=x_{i_j}$ for every $j$. Equivalently, the candidate must equal
  the indexed subsequence $x_I=x_{i_1}\cdots x_{i_m}$. The slimmer may delete
  bases, but it may not substitute, insert, reuse a source position, reorder, or
  invert them.

  The BPNet oracle model expects a fixed input length $L_{\text{min}}$ that is
  larger than $m$ in our evaluations. For any sequence $y$ with
  $|y|\leq L_{\text{min}}$, $\operatorname{Embed}(y,b)$ replaces the centered
  $|y|$-bp region of an $L_{\text{min}}$-bp background $b$ with $y$. When
  $|y|=L_{\text{min}}$, $\operatorname{Embed}(y,b)=y$. All evaluation start
  sequences have $L=1{,}188<L_{\text{min}}$. Let $\energy(z)$ denote the
  oracle energy of input $z$, where lower is better, and let
  $\mathcal{B}_{\mathrm{design}}$ be the set of backgrounds available during
  optimization; it is a singleton in this evaluation. The design problem is
  
  \begin{equation}
  I^* =
  \underset{1\leq i_1<\cdots<i_m\leq L}{\arg\min}
  \frac{1}{|\mathcal{B}_{\mathrm{design}}|}
  \sum_{b\in\mathcal{B}_{\mathrm{design}}}
  \energy\!\left(\operatorname{Embed}(x_{i_1}\cdots x_{i_m},b)\right).
  \label{eq:design}
  \end{equation}

  This two-part output converts the deletion-only requirement into a directly verifiable, executable contract. The evaluator independently reconstructs $x_I$ from the source and rejects the output unless $\widehat{x}=x_I$. Substitutions and insertions fail this reconstruction check. Duplicated, out-of-range, or non-increasing indices fail the index checks, while reordering or inversion fails the index or reconstruction check.

\paragraph{Alternatives and benchmark choices.}
We considered allowing length \emph{at most} $m$ (rather than exactly $m$)
or permitting arbitrary substitutions. Both depart from the intended behavior of a slimmer: at-most length confounds algorithms that return different sizes, and allowing substitutions incentivizes the model to favor them over deletions.  We therefore require exactly $m$ increasing source indices and evaluate several fixed values of $m$ to obtain a length--activity curve. 

\paragraph{Context and activity.}
Rather than pad with ambiguous bases or flank with random sequences, we center each candidate in low-activity backgrounds mined from the human genome.  Design and test
backgrounds are disjoint, and every selected background has lower predicted activity than every embedded start.  For minimized energy $\energy$, held-out background $b$, source $x$, and
candidate $x_I$, the primary metric is
\begin{equation}
R =
\frac{
\energy(b)-\energy\!\left(\operatorname{Embed}(x_I,b)\right)
}{
\energy(b)-\energy\!\left(\operatorname{Embed}(x,b)\right)
}.
\label{eq:retained}
\end{equation}
Thus $R=1$ retains the source's predicted lift over background and $R>1$ improves
it.  Raw candidate energy was rejected as the primary measure because its scale and
baseline vary across starts and contexts.

\paragraph{Compute and uncertainty.}
We follow the NucleoBench protocol and use a wall-clock limit because batching, gradients, and proposal bookkeeping have different costs. Each designed candidate is evaluated in five fixed test backgrounds to measure robustness of the slimmed subsequence. We average the sequence scores across background contexts to obtain one value for each starting sequence, method, and target length. We then compare methods using paired differences on the same starting sequences and bootstrap over starting sequences to obtain nominal confidence intervals.

\section{Agentic discovery of a deletion-only designer}

  \paragraph{Constraining ERA to produce valid DNA slimmers.}
  We ran an ERA program-search task to produce a deletion-only sequence designer with the best $R$. We used hard checks in addition to natural-language instructions. ERA received the task prompt and GrAdaBeam as a starting program, and was asked to modify the designer for deletion-only
  sequence slimming.
  For every proposed program, the evaluator reconstructed the returned
  sequence from those indices, and rejected substitutions, rearrangements, and
  candidates above the requested length. For accepted subsequences, it centered each variable-length sequence in one target-specific low-activity background before scoring it. The instructions prioritized reaching the length constraint before optimizing activity. Together,
  these choices prevented ERA from ``cheating'' the deletion-only constraint, while leaving it free to explore algorithm space.

  \paragraph{ERA program search.}
Each proposed designer was evaluated on one high-scoring ATAC starting sequence and one high-scoring MYC starting sequence. For each case, the designer optimized the corresponding BPNet-lite oracle using that target's low-activity background. The returned sequence was scored with the same target-specific oracle, and the two resulting energies were averaged to produce the scalar
objective supplied to ERA. Failures, timeouts, and invalid subsequences received the worst objective value, while successful programs were expanded through ERA's program-search procedure.

\paragraph{The discovered slimmer.}
The best program, \method{}, maintains a salience value for
each source position.  Above the target length, it repeatedly removes approximately half of the remaining length gap, sampling deletions preferentially from low-salience positions and
keeping the best of 64 proposals.  At exact length, it proposes membership swaps,
single-index shifts, and contiguous block shifts.  It updates salience by crediting
the positions retained in better-than-average candidates and diffuses a small
amount of credit to neighboring positions.  A cooling simulated-annealing rule
occasionally accepts worse exact-length states, while an archive preserves the
best valid candidate \citep{vanlaarhoven1987annealing}. ERA's resulting program
combines rapid deletion, exact-length refinement, and persistent empirical
salience learned from scored candidate batches. The
search and constants are reported in
Appendix~\ref{app:algorithm}.

\section{Five-target BPNet evaluation}

\paragraph{Protocol.}
The evaluation consists of slimming tasks for transcription factors E2F3, ELF4, MAX, MECOM, and RAD21, none of
which were used during ERA program search. Each study compares random exact
deletion, batched greedy deletion with exact-length swap refinement, and ERA
\method{} on five high-activity 1,188-bp starts at target lengths 400 and
100 bp. Every evaluation run receives a 600-second wall-clock budget on an H200 GPU to produce the best target-length sequence. Each target has one fixed target-specific design
context visible during optimization. Thus, each study contains 30 unique design cells and 150
background-level evaluation rows. The five test-context scores are averaged first, yielding one
value per start, method, and length. Method differences are paired by start,
and bootstrap intervals resample the five starts. Each target and length is
reported separately.

\paragraph{Results.}

\begin{table}[t]
  \centering
  \scriptsize
  \setlength{\tabcolsep}{2.5pt}
  \caption{Retained effect (higher is better). Means average five starts after
  averaging five test contexts within each start. ERA$-$greedy intervals and
  wins are paired by start.}
  \label{tab:heldout}
  \begin{tabular}{llrrrrr}
    \toprule
    Target & bp & Random & Greedy & ERA & ERA$-$greedy [95\% CI] & Wins/5 \\
    \midrule
    E2F3 & 400 & 0.862 & 3.918 & 5.735 & 1.817 [1.320, 2.321] & 5/5 \\
    E2F3 & 100 & 0.613 & 3.012 & 2.713 & -0.299 [-0.516, -0.072] & 1/5 \\
    ELF4 & 400 & 0.636 & 1.533 & 1.999 & 0.466 [0.398, 0.533] & 5/5 \\
    ELF4 & 100 & 0.348 & 1.083 & 1.152 & 0.069 [-0.009, 0.129] & 4/5 \\
    MAX & 400 & 1.068 & 1.810 & 2.211 & 0.401 [0.321, 0.484] & 5/5 \\
    MAX & 100 & 0.748 & 1.448 & 1.519 & 0.071 [-0.027, 0.170] & 3/5 \\
    MECOM & 400 & 0.645 & 2.474 & 4.065 & 1.590 [1.412, 1.785] & 5/5 \\
    MECOM & 100 & 0.385 & 1.781 & 1.919 & 0.138 [-0.026, 0.322] & 3/5 \\
    RAD21 & 400 & 0.475 & 1.148 & 1.754 & 0.606 [0.365, 0.848] & 5/5 \\
    RAD21 & 100 & 0.280 & 0.955 & 1.142 & 0.187 [-0.008, 0.367] & 4/5 \\
    \bottomrule
  \end{tabular}
\end{table}

ERA has the highest mean retained effect in nine of ten settings and exceeds
random deletion in all ten. At 400 bp, ERA outperforms greedy for every target
and all 25 paired starts, with every target-specific interval above zero. At
100 bp, ERA is numerically higher for ELF4, MAX, MECOM, and RAD21, but all four
intervals overlap zero. Note that this regime is significantly shorter than what ERA used during program search (ERA focused on slimming to 300 bp). Greedy outperforms ERA for E2F3 at 100 bp (mean
difference $-0.299$, 95\% CI $[-0.516,-0.072]$). ERA's mean retained effect
exceeds 1 in every setting, indicating that its slimmed sequences improve the predicted lift over background relative to their full-length sources.

\section{Limitations and conclusions}

The evaluation design has five starts and one optimizer seed per target and tests
only two output lengths. Targets were selected using results from NucleoBench, and the starts were selected for high activity. Contexts
are fixed and target-specific. The same target oracle is used for optimization
and evaluation, so the study does not test an orthogonal model; there is also
no wet-lab validation. ERA program search did not include 100-bp targets, so the
100-bp evaluation tests substantially more aggressive slimming than was used
during algorithm discovery. The frequent \(R>1\) values, including for 100-bp sequences, may reflect exploitation of the BPNet oracle or artificial deletion junctions rather than preservation of biological function. Bootstrap intervals are based on only five starting sequences per target and therefore provide limited-resolution uncertainty estimates.

Together, this work establishes deletion-only slimming as a distinct, machine-verifiable design problem. The preliminary results show a consistent ERA advantage at 400 bp and mixed performance at 100 bp.

\bibliographystyle{plainnat}
\bibliography{agentls_gradaslim_short}

\appendix

\section{Complete discovered algorithm}
\label{app:algorithm}

Algorithm~\ref{alg:gradaslim} gives the exact-length evaluation implementation
of \method{}. ERA discovered its proposal and credit-assignment rules.

\refstepcounter{algorithm}
\noindent\textbf{Algorithm \thealgorithm: ERA-discovered \method{}}\label{alg:gradaslim}
\begin{algorithmic}[1]
\Require source $x$ of length $L$; target $m$; minimized oracle $f$; budget $T$;
seed $q$; batch size $B=64$
\State $t_0\gets\Call{Clock}{}$
\State initialize isolated RNG with $q$
\State $s_i\sim\mathcal{N}(0,10^{-5})$ for $i=1,\ldots,L$
\State score exact fallback $F=(1,\ldots,m)$; $(I^*,e^*)\gets(F,f(x_F))$
\State $I\gets(1,\ldots,L)$; $e\gets f(x_I)$
\While{$\Call{Clock}{}-t_0<T$}
  \If{$|I|>m$} \Comment{Phase 1: reach the constraint quickly}
    \State $d\gets\max\{1,\lceil(|I|-m)/2\rceil\}$
    \For{$j=1,\ldots,B$}
      \State $w_k\gets \max_{r\in I}s_r-s_k+0.1\,\mathrm{sd}(s_I)+10^{-7}$
      \State sample $d$ distinct positions $D_j\subset I$ proportional to $w$
      \State $J_j\gets I\setminus D_j$
    \EndFor
  \Else \Comment{Phase 2: exact-length refinement}
    \State $U\gets\{1,\ldots,L\}\setminus I$
    \For{$j=1,\ldots,B$}
      \State $J_j\gets I$; draw $r\sim\mathrm{Uniform}(0,1)$
      \If{$r<0.35$ and $U\neq\emptyset$} \Comment{salience-biased swap}
        \State sample $a\in I$ proportional to $\max(s_I)-s_a+10^{-7}$
        \State sample $u\in U$ proportional to $s_u-\min(s_U)+10^{-7}$
        \State $J_j\gets\mathrm{sort}((I\setminus\{a\})\cup\{u\})$
      \ElsIf{$r<0.65$} \Comment{point shift}
        \State sample retained rank $k$ uniformly
        \State replace $I_k$ uniformly within its order-preserving neighbor bounds
      \Else \Comment{contiguous block shift}
        \State sample a valid block start and a block length $\ell\in\{1,\ldots,11\}$
        \State shift the block uniformly within its order-preserving bounds
      \EndIf
    \EndFor
  \EndIf
  \State $e_j\gets f(x_{J_j})$ for $j=1,\ldots,B$ in one batch;
         $\bar e\gets B^{-1}\sum_j e_j$
  \State $e^{*,\mathrm{old}}\gets e^*$ \Comment{archive best before this iteration}
  \For{$j=1,\ldots,B$ and $k\in J_j$}
    \State $s_k\gets s_k+(\bar e-e_j)$ \Comment{credit bases in good candidates}
  \EndFor
  \For{$k=2,\ldots,L-1$}
    \State $s_k\gets s_k+0.05(s_{k-1}+s_{k+1})$ \Comment{neighbor diffusion}
  \EndFor
  \State $j^*\gets\arg\min_j e_j$
  \If{$|I|>m$}
    \State $(I,e)\gets(J_{j^*},e_{j^*})$
  \Else
    \State $p\gets\min\{1,(\Call{Clock}{}-t_0)/T\}$;
           $\tau\gets\max\{0.001,0.4(1-p)^3\}$
    \If{$e_{j^*}<e$ or $\mathrm{Uniform}(0,1)<\exp(-(e_{j^*}-e)/\tau)$}
      \State $(I,e)\gets(J_{j^*},e_{j^*})$
    \EndIf
  \EndIf
  \State if any $|J_j|=m$ and $e_j<e^*$, set $(I^*,e^*)$ to the best such
         $(J_j,e_j)$
  \If{$|I|=m$ and $e<e^{*,\mathrm{old}}$} \Comment{reinforce a newly accepted global best}
    \State $s_I\gets s_I+0.1|e|$
  \EndIf
\EndWhile
\State \Return $(x_{I^*},I^*)$
\end{algorithmic}

\paragraph{Interpretation.}
The salience vector is a persistent credit-assignment reservoir over source
coordinates, not a model gradient.  For each scored batch, every retained index
receives reward $\bar e-e_j$: positions occurring in below-average-energy
candidates gain salience.  Phase 1 uses the inverse of this value to target
deletions.  Phase 2 removes low-salience retained positions and restores
high-salience omitted positions, while point and block shifts search local source
geometry.  Neighbor diffusion biases nearby bases together, which may preserve
motifs but may also amplify correlated artifacts.  Simulated annealing allows the
single incumbent to cross local barriers, whereas the archive prevents loss of the
best exact solution.

\end{document}